\documentclass[10pt,twocolumn,a4paper]{esaAI}
\usepackage{booktabs}
\usepackage{array}
\usepackage{xcolor}
\usepackage{marvosym}

\title{Low-power Ship Detection in Satellite Images Using Neuromorphic Hardware}

\author[1]{Gregor Lenz\thanks{Corresponding author. E-Mail: gregor@neurobus.space}}
\author[2]{Douglas McLelland}
\affil[1]{Neurobus, Toulouse, France}
\affil[2]{BrainChip, Toulouse, France}

\begin{document}

\makeCustomtitle

\begin{abstract}

Transmitting Earth observation image data from satellites to ground stations incurs significant costs in terms of power and bandwidth. For maritime ship detection, on-board data processing can identify ships and reduce the amount of data sent to the ground. However, most images captured on board contain only bodies of water or land, with the Airbus Ship Detection dataset showing only 22.1\% of images containing ships.
We designed a low-power, two-stage system to optimize performance instead of relying on a single complex model. The first stage is a lightweight binary classifier that acts as a gating mechanism to detect the presence of ships. This stage runs on Brainchip's Akida 1.0, which leverages activation sparsity to minimize dynamic power consumption. The second stage employs a YOLOv5 object detection model to identify the location and size of ships. This approach achieves a mean Average Precision (mAP) of 76.9\%, which increases to 79.3\% when evaluated solely on images containing ships, by reducing false positives.
Additionally, we calculated that evaluating the full validation set on a NVIDIA Jetson Nano device requires 111.4 kJ of energy. Our two-stage system reduces this energy consumption to 27.3 kJ, which is less than a fourth, demonstrating the efficiency of a heterogeneous computing system.

\end{abstract}

\section{Introduction}
Ship detection from satellite imagery is a critical application within the field of remote sensing, offering significant benefits for maritime safety, traffic monitoring, and environmental protection. 
The vast amount of data generated by satellite imagery cannot all be treated on the ground in data centers, as the downlinking of image data from a satellite is a costly process in terms of power and bandwidth. 
\begin{figure}[t]
    \centering
    \includegraphics[width=\columnwidth]{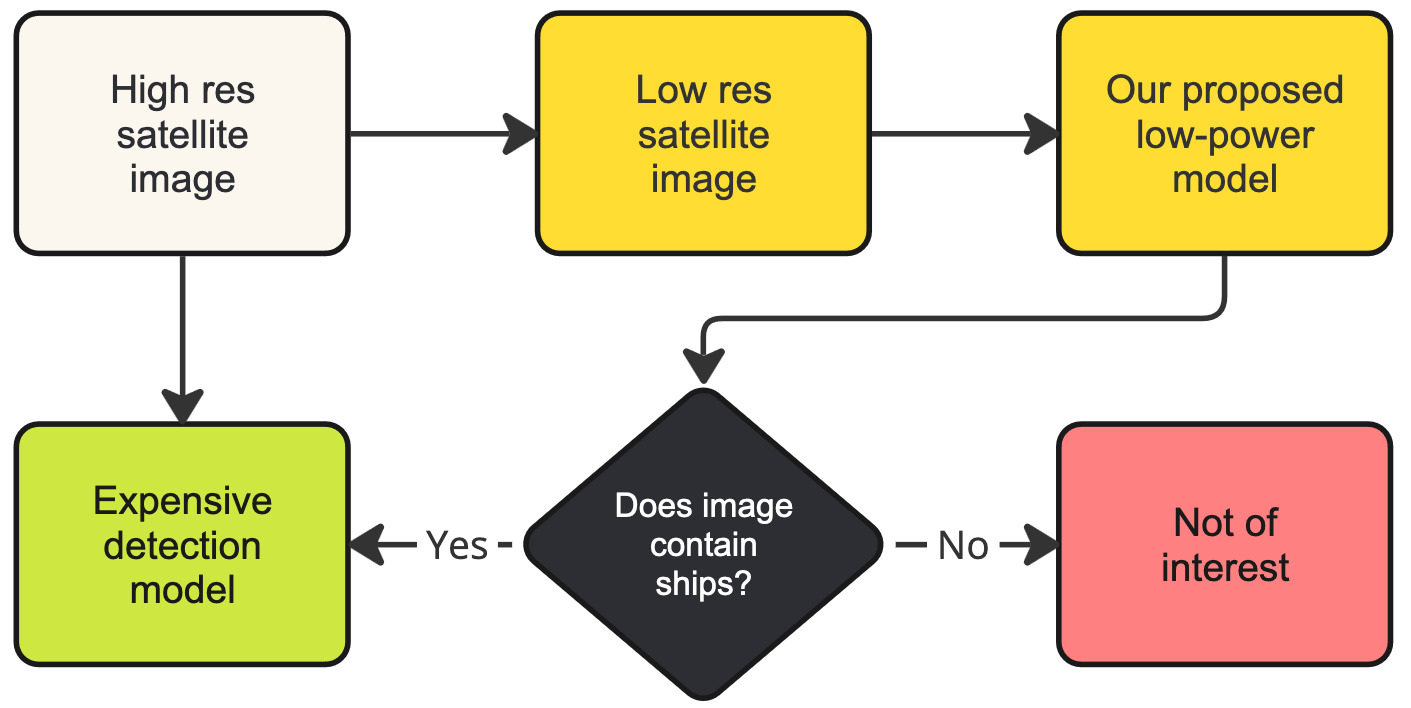}
    \caption{Data flow chart of our system.}
    \label{fig:model-design}
\end{figure}
To help  satellites identify the most relevant data to downlink and alleviate processing on the ground, recent years have seen the emergence of edge artificial intelligence (AI) applications for Earth observation~\cite{xu2022lite, zhang2020ls, xu2021board, ghosh2021board, yao2019board, alghazo2021maritime, vstepec2019automated}. By sifting through the data on-board the satellite, we can discard a large number of irrelevant images and focus on the relevant information.
Because satellites are subject to extreme constraints in size, weight and power, energy-efficient AI systems are crucial. In response to these demands, our research focuses on using low-power neuromorphic chips for ship detection tasks in satellite images. Neuromorphic computing, inspired by the neural structure of the human brain, offers a promising avenue for processing data with remarkable energy efficiency. 
The Airbus Ship Detection challenge~\cite{al2021airbus} on Kaggle aimed to identify the best object detection models. A post-challenge analysis~\cite{faudi2023detecting} revealed that a binary classification pre-processing stage was crucial in winning the challenge, as it reduced the rates of false positives and therefore boosted the relevant segmentation score.  
We introduce a ship detection system that combines a binary classifier with a powerful downstream object detection model. The first stage is implemented on a state-of-the-art neuromorphic chip and determines the presence of ships. Images identified as containing ships are then processed by a more complex detection model in the second stage, which can be run on more flexible hardware.
Our work showcases a heterogeneous computing pipeline for a complex real-world task, combining the low-power efficiency of neuromorphic computing with the increased accuracy of a more complex model.


\begin{figure*}[ht]
    \centering
    \includegraphics[width=\textwidth]{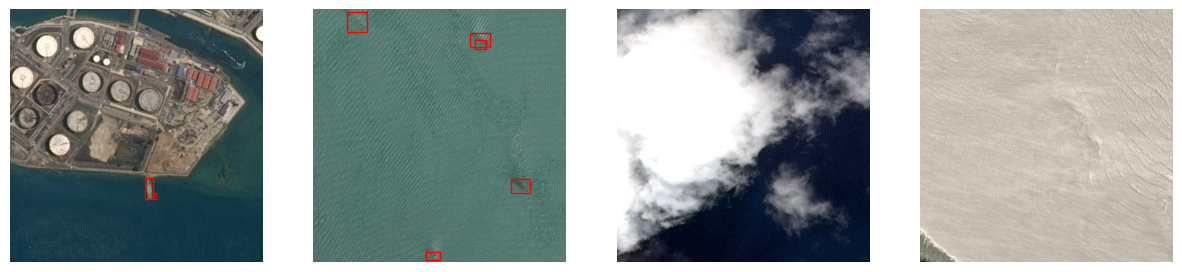}
    \caption{The first two images are examples of 22\% of annotated samples. The second two images are examples of the majority of images that do not contain ships but only clouds, water or land.}
    \label{fig:sample-images}
\end{figure*}

\section{Dataset}
The Airbus Ship Detection dataset~\cite{airbus_ship_detection_2018} contains 192k satellite images, of which 22.1\% contain annotated bounding boxes for a single \emph{ship} class. Key metrics of the dataset are described in \cref{tab:dataset}.  As can be seen in the sample images in \cref{fig:sample-images}, a large part of the overall pixel space captures relatively homogenuous parts such as open water or clouds. 
We chose this dataset as it is part of the European Space Agency's (ESA) On-Board Processing Benchmark suite for machine learning applications~\cite{obpmark}, with the goal in mind to test and compare a variety of edge computing hardware platforms for the most common ML tasks related to space applications. 
The annotated ship bounding boxes have diagonals that vary from 1 to 380 pixels in length, and 48.3\% of bounding boxes have diagonals of 40 pixels or shorter. Given that the images are $768 \times 768$px in size, this makes it a challenging dataset, as the model needs to be able to detect ships of a large variety of sizes. 
Since on Kaggle there are only annotations for the training set available, we used a random 80/20 split for training and validation, similarly to Huang et al~\cite{huang2020fast}.
For our binary classifier, we downsized all images to $256 \times 256$px, to be compatible with the input resolution of Akida 1.0, and labeled the images as 1 if they contained at least one bounding box of any size, otherwise 0.  For our detection model, we downsized all images to $640 \times 640$px in size.


\begin{table}[htb]
    \centering
    \resizebox{\columnwidth}{!}{ 
    \begin{tabular}{lc}
        \toprule
        RGB image size & $768 \times 768$ \\
        Total number of images & 192,555 \\
        Number of training images & 154,044 \\
        Percentage of images that contain ships & 22.1\% \\
        Total number of bounding boxes & 81,723 \\
        Median diagonal of all bounding boxes & 43.19px \\
        Ratio of bounding box to image area & 0.3\% \\
        \bottomrule
    \end{tabular}
    }
    \caption{Summary of image and bounding box data for the Airbus Ship Detection Training dataset.}
    \label{tab:dataset}
\end{table}

\section{Models}
For our binary classifier, we used a 866k parameter model named \emph{AkidaNet 0.5}, which is loosely inspired from \mbox{MobileNet} \cite{howard2017mobilenets} with alpha = 0.5. It consists of standard convolutional, separable convolutional and linear layers, to reduce the number of parameters and to be compatible with Akida 1.0 hardware. To train the network, we used binary crossentropy loss, the Adam optimizer, a cosine decay learning rate scheduler with initial rate of 0.001 and lightweight L1 regularization on all model parameters over 10 epochs. 
For our detection model, we trained a YOLOv5 medium~\cite{ge2021yolox} model of 25m parameters with stochastic gradient descent, a learning rate of 0.01 and 0.9 momentum, plus blurring and contrast augmentations over 25 epochs. 

\section{Akida hardware}
\begin{figure}[ht]
    \centering
    \includegraphics[width=\columnwidth]{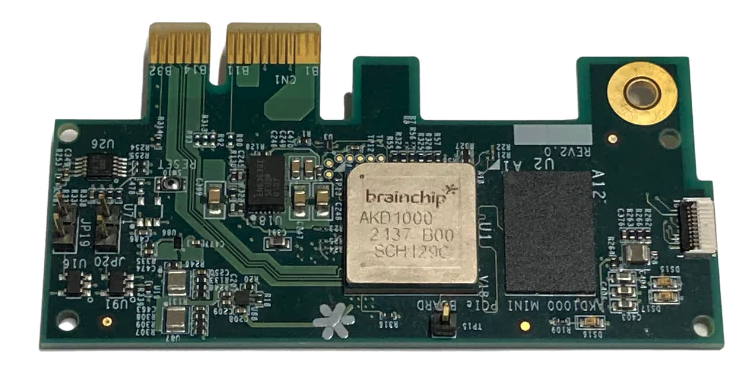}
    \caption{AKD1000 chip on a PCB with PCIe x1 connector.}
    \label{fig:akida}
\end{figure}

Akida by Brainchip is an advanced artificial intelligence processor inspired by the neural architecture of the human brain, designed to provide high-performance AI capabilities at the edge with exceptional energy efficiency. Version 1.0 is available for purchase in the form factor of PCIe x1 as shown in \cref{fig:akida}, and supports convolutional neural network architectures. Version 2.0 adds support for a variety of neural network types including RNNs and transformer architectures, but is currently only available in simulation. The Akida processor operates in an event-based mode for intermediate layer activations, which only performs computations for non-zero inputs, significantly reducing operation counts and allowing direct, CPU-free communication between nodes. Akida 1.0 supports flexible activation and weight quantization schemes of 1, 2, or 4 bit. Models are trained in Brainchip's MetaTF, which is a lightweight wrapper around Tensorflow. In March 2024, Akida has also been sent to space for the first time~\cite{brainchip2024launch}. 

\section{Results}
\subsection{Classification accuracy}
The key metrics for our binary classification model are provided in \cref{tab:model-metrics}. The trained floating point model reaches an accuracy of 97.91\%, which drops to 95.75\% after quantizing the weights and activations to 4 bit and the input layer weights to 8 bit. After one epoch of quantization-aware training with a tenth of the normal learning rate, the model recovers nearly its floating point accuracy, at 97.67\%. Work by Alghazo et al~\cite{alghazo2021maritime} reaches an accuracy of 89.7\% in the same binary classification setting, albeit on a subset of the dataset and on images that are downscaled to 100 pixels. In addition, the corresponding recall and precision metrics for our model are shown in the table. In our system we prioritize recall, because false negatives (missing a ship) have a higher cost than false positives (detecting ships where there are none), as the downstream detection model can correct for mistakes of the classifier stage. By default we obtain a recall of 94.4 and a precision of 95.07\%, but by adjusting the decision threshold on the output, we bias the model to include more images at the cost of precision, obtaining a recall of 97.64\% for a precision of 89.73\%. 

\begin{table}[ht]
    \centering
    \caption{Model performance comparison in percent. FP is floating point, 4 bit is the quantized model of 8 bit inputs, and 4 bit activations and weights. QAT is quantization-aware training for 1 epoch with reduced learning rate. Precision and recall values are given for a decision threshold of 0.5 and 0.1.}
    \begin{tabular}{lccc}
        \toprule
         & FP & 4 bit & After QAT \\
        \midrule
        Accuracy  & 97.91 & 95.75 & \textbf{97.67} \\
        Accuracy \cite{alghazo2021maritime} & 89.70 & -     & -     \\
        Recall    & 95.23 & 85.12 & 94.40/\textbf{97.64} \\
        Precision & 95.38 & 95.32 & 95.07/\textbf{89.73} \\
        \bottomrule
    \end{tabular}
    \label{tab:model-metrics}
\end{table}

\subsection{Performance on Akida 1.0}
The model that underwent QAT is then deployed to the Akida 1.0 reference chip, AKD1000, where the same accuracy, recall and precision are observed as in simulation. As detailed in \cref{tab:hardware-metrics}, feeding a batch of 100 input images takes 1.168\,s and consumes 440\,mW of dynamic power. The dynamic energy used to process the whole batch is therefore 515\,mJ, which translates to 5.15\,mJ per image. The network is distributed across 51 of the 78 available neuromorphic processing cores. During our experiments, we measured 921\,mW of static power usage on the AKD1000 reference chip. We note that this value is considerably reduced in later chip generations.

\begin{table}[ht]
\centering
\caption{Summary of performance metrics on Akida 1.0 for a batch size of 100. }
\begin{tabular}{lc}
\toprule
Total duration (ms)               & 1167.95 \\
Duration per sample (ms)          & 11.7 \\
Throughput (fps)                  & 85.7 \\
Total dynamic power (mW)                  & 440.8 \\
Energy per batch (mJ)             & 514.84 \\
Energy per sample (mJ)            & 5.15 \\
Total neuromorphic processing cores & 51     \\
\bottomrule
\end{tabular}
\label{tab:hardware-metrics}
\end{table}

\begin{figure}[htb]
    \centering
    \includegraphics[width=\columnwidth]{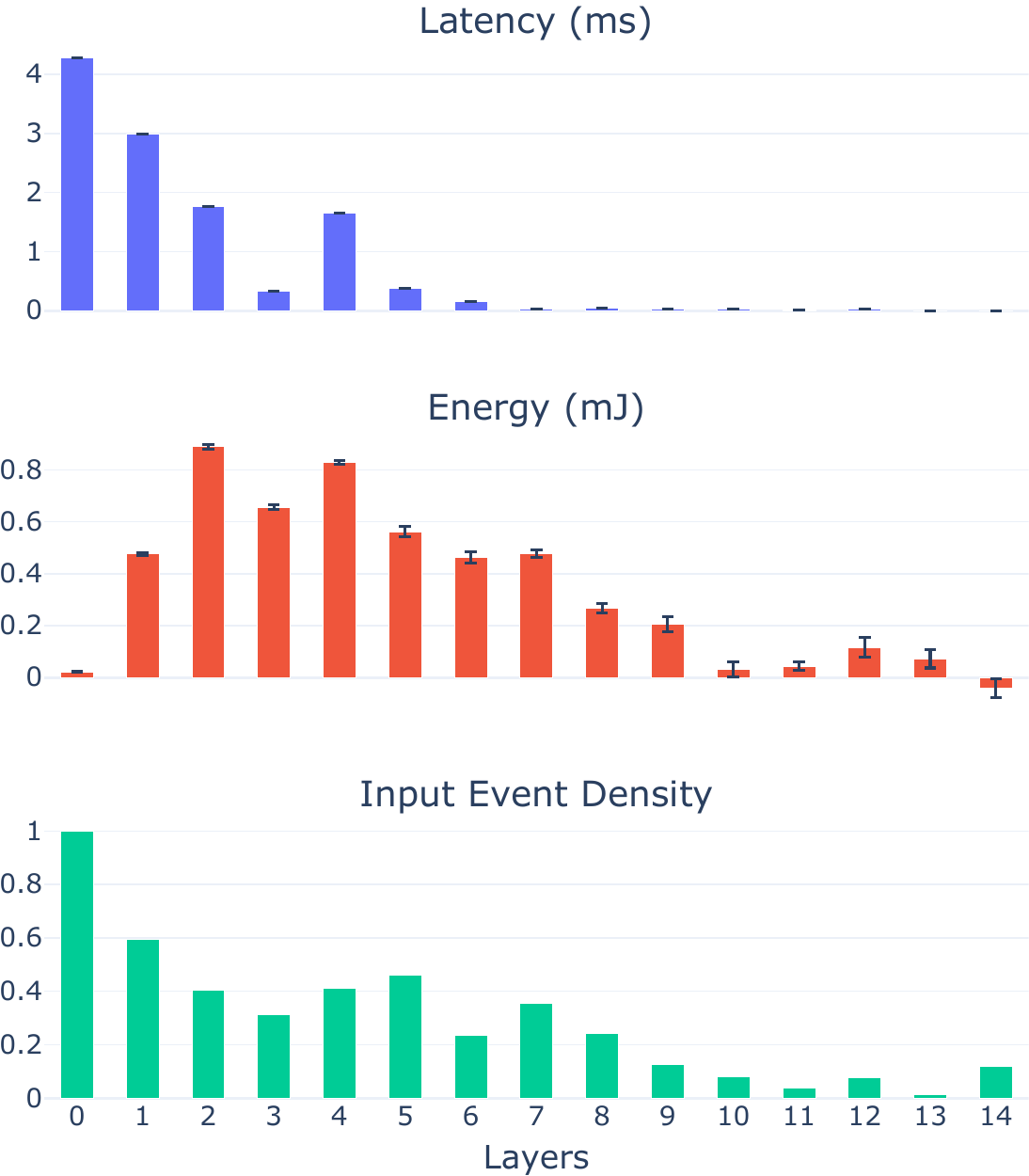}
    \caption{Layer-wise statistics per sample image for inference of binary classifier on Akida v1, measured for a batch of 100 images over 10 repeats. }
    \label{fig:layer-stats}
\end{figure}

We can further break down performance across the different layers in the model. 
The top plot in \cref{fig:layer-stats} shows the latency per frame: it increases as layers are added up to layer 7, but beyond that, the later layers make almost no difference. 
As each layer is added, we can measure energy consumption, and estimate the per-layer contribution as the difference from the previous measurement, shown in the middle plot of \cref{fig:layer-stats}. 
We observe that most of the energy during inference is spent on earlier layers, even though the work required per layer is expected to be relatively constant as spatial input sizes are halved, but the number of filters doubled throughout the model.
The very low energy measurements of later layers are explained by the fact that Akida is an event-based processor that exploits sparsity. When measuring the density of input events per layer as shown in the bottom plot of \cref{fig:layer-stats}, we observe that energy per layer correlates well with the event density. The very first layer receives dense images, but subsequent layers have much sparser inputs, presumably due to a lot of input pixels that are not ships, which in turn reduces the activation of filters that encode ship features. We observe an average event density of 29.3\% over all layers including input, reaching less than 5\% in later layers. This level of sparsity is achieved  through the combination of ReLU activation functions and L1 regularization on activations during training.


\subsection{Detection model performance}
For our subsequent stage, our YOLOv5 model of 25m parameters achieves 76.9\% mAP when evaluated on the full validation set containing both ship and non-ship images. When we evaluate the same model on the subset of the validation set that just contains ships, the  mAP jumps to 79.3\%, as the false positives are reduced considerably. That means that our classifier stage already has a beneficial influence on the detection performance of the downstream model. 
\cref{tab:detection-results} provides an overview of detection performance in the literature. 
Machado et al.~\cite{machado2022estimating} provide measurements for different YOLOv5 models on the NVIDIA Jetson Nano series, a hardware platform designed for edge computing. For the YOLOv5 medium model, the authors report an energy consumption of 0.804 mWh per frame and a throughput of 2.7 frames per second at input resolution of 640 pixels, which translates to a power consumption of 7.81\,W. The energy necessary to process the full validation dataset of 38,511 images on a Jetson is therefore $ 38511 \times 7.81 / 2.7 = 111.4$ kJ. 
For our proposed two-stage system, we calculate the total energy as the sum of processing the full validation set on Akida plus processing the identified ship images on the Jetson device. Akida has a power consumption of $0.921 + 0.44 = 1.361$\,W at a throughput of 85.7 images/s.
With a recall of 97.64\% and a precision of 89.73\%, 9243 images, equal to 24.03\% of the validation data, are classified to contain ships, in contrast to the actual 22.1\%.
We therefore obtain an overall energy consumption of 
$38511 \times 1.361 / 85.7 + 9243 \times 7.81 / 2.7 = 27.3$ kJ.
Our proposed system uses 4.07 times less energy to evaluate this specific dataset. 


\begin{table}[htb]
    \centering
    \begin{tabular}{ccc}
        \toprule
        Model & mAP (\%)& Energy (kJ)\\
        \midrule
       YOLOv3~\cite{patel2022deep} & 49 & - \\
       YOLOv4~\cite{patel2022deep} & 61 & - \\
       YOLOv5~\cite{patel2022deep} & 65 & - \\
       Faster RCNN~\cite{al2021airbus} & 80 & - \\
       \midrule
YOLOv5 & 76.9 & 111.4 \\
AkidaNet + YOLOv5 & 79.3 & 27.3 \\
        \bottomrule
    \end{tabular}
    \caption{Mean average precision and energy consumption evaluated on the Airbus ship detection dataset.}
    \label{tab:detection-results}
\end{table}

\section{Discussion}
For edge computing tasks, it is common to have a small \emph{gating} model which activates more costly downstream processing whenever necessary. As only 22.1\% of images in the Airbus detection dataset contain ships, a two-stage processing pipeline can leverage different model and hardware architectures to optimize the overall system. 
We show that our classifier stage running on Akida benefits from a high degree of sparsity when processing the vast amounts of homogeneous bodies of water, clouds or land in satellite images, where only 0.3\% of the pixels are objects of interest. We hypothesise that many filter maps that encode ship features are not activated most of the times. This has a direct impact on the dynamic power consumption and latency during inference due to Akida's event-based processing nature.
In addition, we show that a two-stage system actually increases the mAP of the downstream model by reducing false positive rates, as is also mentioned in the post-challenge analysis of the Airbus Kaggle challenge~\cite{faudi2023detecting}. 
The energy consumption of the hybrid system is less than a fourth compared to running the detection model on the full dataset, with more room for improvement when using Akida v2, which is going to reduce both static and dynamic power consumption and allow the deployment of more complex models that likely achieve higher recall rates. 
The limitations of our system are the increased needs of size, having to fit two different accelerators instead of a single one. But by combining the strengths of different hardware platforms, we can optimize the overall performance, which is critical for edge computing applications in space. 

\printbibliography
\addcontentsline{toc}{section}{References}


\end{document}